\documentclass{article}
\usepackage{ijcai20}

\usepackage{times}
\usepackage{soul}
\usepackage{url}
\usepackage[hidelinks]{hyperref}
\usepackage[utf8]{inputenc}
\usepackage[small]{caption}
\usepackage{graphicx}
\usepackage{amsmath}
\usepackage{amsmath,bm}
\usepackage{amsfonts,amssymb}
\usepackage{booktabs}
\usepackage{algorithm}
\usepackage{algorithmic}
\usepackage{subfigure}
\title{Learning Relation Ties with a Force-Directed Graph in Distant Supervised Relation Extraction}

\author{
    Yu-Ming Shang, He-Yan Huang, Xin Sun, Xian-Ling Mao
}

\author{
Yuming Shang
\and
Heyan Huang\and
Xin Sun\And
Xianling Mao
\affiliations
Beijing Institute of Technology
\emails
ymshang@bit.edu.cn
}

\newcommand{\citet}[1]{\citeauthor{#1} \shortcite{#1}}
\newcommand{\citep}{\cite}

\begin{document}

\maketitle

\begin{abstract}
	% Relation ties的定义，这一句不改了
	Relation ties, defined as the correlation and mutual exclusion between different relations, are critical for distant supervised relation extraction.
	% 以往方法的做法
	Existing approaches model this property by greedily learning local dependencies.
	% 描述问题
	However, they are essentially limited by failing to capture the global topology structure of relation ties. 
	As a result, they may easily fall into a locally optimal solution.
	% 为了解决这一模型，我们借鉴了库仑力
	To solve this problem, in this paper, we propose a novel force-directed graph based relation extraction model to comprehensively learn relation ties.
	% 具体来说
	Specifically,
	% 我们首先将关系之间的联系构建成一张图
	we first build a graph according to the global co-occurrence of relations.
	% 然后，我们借用
	Then, we borrow the idea of Coulomb's Law from physics and introduce the concept of attractive force and repulsive force to this graph to learn correlation and mutual exclusion between relations.
	% 最后
	Finally, the obtained relation representations are applied as an inter-dependent relation classifier.
	% 实验证明
	Experimental results on a large scale benchmark dataset demonstrate that our model is capable of modeling global relation ties and significantly outperforms other baselines. 
	Furthermore, the proposed force-directed graph can be used as a module to augment existing relation extraction systems and improve their performance.
	% 第二个层面
	%and can indeed acquire the implicit connections between relations. 

\end{abstract}

\section{Introduction}
	% Distant Supervised Relation Extracion
	Relation extraction, defined as the task of extracting structured relations from primitive unstructured text, is crucial in natural language processing (NLP). 
	% 开始说远程监督以及远程监督中出现的问题
	Conventional supervised methods are time-consuming for the requirement of large-scale manually labeled data.
	Therefore, ~\citet{mintz2009distant} propose distant supervision to automatically label sentences. It assumes that if two entities have a relation $r$ in a knowledge graph, then any sentence that mentions the two entities might express that relation.
	As there may be multiple relations between one entity pair, distant supervised relation extraction is a multi-label prediction task.
	
	\begin{figure}[t]
		\centering
		\includegraphics[width=0.99\columnwidth]{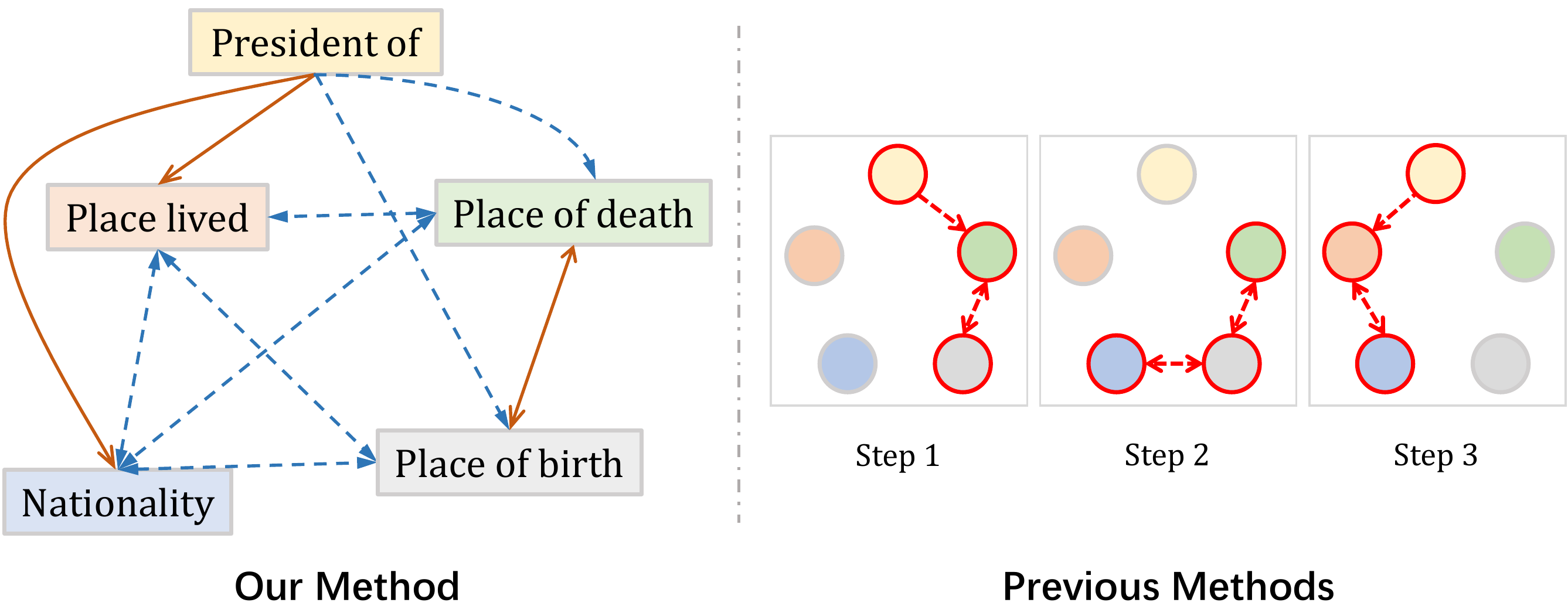} 
		\caption{The difference between our method and previous methods when learning relation ties. The solid arrow indicates that the new relation must exist. The dashed arrow indicates that the new relation may exist.}
		\label{fig:example}
		\vspace{-1.5pt}
	\end{figure}

	% 引出relation ties
	Interestingly, relations in distant supervision usually have inner correlation and mutual exclusion, which we call \textit{relation ties}.
	As shown in Figure \ref{fig:example}, 
	% 描述相似性
	if an entity pair (\textsl{Barack Obama}, \textsl{United States}) has a relation \texttt{President\_of}, then we can infer that the entity pair must have relations \texttt{Place\_lived} and \texttt{Nationality}. Similarly, the relation \texttt{Place\_of\_birth} may also exist between the two entities in a certain probability. 
	% 描述互斥性
	On the contrary, we can also infer that the entity pair does not contain relation \texttt{Location}. 
	Because the head entity \textsl{Barack Obama} is the name of a person, not a place.
	% 说Relation Ties的作用
	Obviously, considering relation ties can effectively narrow down the potential searching space and significantly improve the performance of relation extraction.
	
	% 有很多方法研究这个问题
	Existing studies on learning relation ties broadly fall into two types: the explicit methods and the implicit methods. 
	% 介绍显示模型
	The former tries to use model architecture to explicitly represent dependencies and conflicts between relations, e.g., Markov Logic Network \cite{han2016global} or Encoder-Decoder framework \cite{su2018exploring}.
	% 介绍隐约的模型
	The latter implicitly learns relation ties by soft constraints, e.g., designing loss function \cite{jiang2016relation,jointly} or using attention mechanism \cite{feng2017effective}.
	% 这句话还是要研究一下这个论文再决定该咋说，今晚要搞定摘要
	However, as shown in the right part of Figure \ref{fig:example}, previous approaches greedily obtain local dependencies between relations at each learning step, and have difficulty in making a global optimization. 
	% 说存在的问题
	As a consequence, they fail to precisely describe the complex global topology structure of relation ties and may easily fall into a locally optimal solution.
	
	% 这一段还是需要好好地斟酌一下呀，尴尬了
	To address this issue, in this paper, we propose a novel \textbf{F}orce-\textbf{D}irected \textbf{G}raph based \textbf{R}elation \textbf{E}xtraction model, named FDG-RE, which is able to comprehensively learn the global relation ties in an end-to-end manner.
	Specifically,
	% 怎么构建这个图很关键
	we build a graph based on global co-occurrence of relations, where each node is represented by a relation embedding and the edge indicates the co-occurrence between two relations. 
	% 说这个Graph理想的形状
	Intuitively, the ideal topology structure of the graph should be that related nodes are close while conflicted nodes are far away. 
	% 我们定义了两种力
	To this end, we borrow the Coulomb's Law \cite{halliday2013fundamentals} from physics and introduce the concept of attractive force and repulsive force into the graph. 
	% 力的作用 
	The aim of attractive force is to increase the similarity between two correlated relation embeddings so that they are close with each other in the embedding space, while the repulsive force works in the opposite direction.
	% 为了增加引力
	To simulate attractive force, we employ graph convolutional network (GCN) \cite{kipf2017semi} to obtain information propagation between correlated relation embeddings.  
	% 为了增加斥力
	To simulate repulsive force, we utilize the similarity between conflicted relation embeddings as a penalty term for the objective loss function.
	% 分类器
	Finally, the relation representations learned by the force-directed graph are applied as an inter-dependent relation classifier. 
	% 实验结果
	Experimental results prove that our FDG-RE performs better than the state-of-the-art baselines. 
	
	% 本文的贡献
	To sum up, our contributions can be encapsulated as follows:
	
	\begin{itemize}
		
		\item Different from existing methods, the proposed FDG-RE can precisely learn global topology structure of relation ties.
		
		\item The proposed force-directed graph can be applied as an independent module to other relation extraction methods to improve their performance.
		
		\item Experiments on a widely used dataset prove that our FDG-RE achieves state-of-the-art performance.
	\end{itemize}
	
	\section{Motivations}
	
	In distant supervision scenario, every relation has its own correlated relations and conflicted relations. 
	Therefore, if we want to precisely learn global relation ties, the following three core problems must be solved:
	
	(1) What kind of data structure is appropriate for describing relation ties?
	
	Intuitively, the correlation and mutual exclusion between relations constitute a complex network. In order to comprehensively represent all connections in this network, we build a graph based on the co-occurrence of relations. In this graph, each node is represented by a relation embedding and the edge between two nodes indicates the co-occurrence dependency of the two relations. 
	
	(2) What is the ideal topology structure of the graph?
	
	To explore this issue, we borrow the idea of Coulomb's Law from physics. The law states that the force between two charges ($q_1, q_2$) is directly proportional to the product of them and inversely to the square of the distance between them. 
	If $q_1 \cdot q_2 >0$, the force is negative (repulsive force), if $q_1 \cdot q_2 <0$, the force is positive (attractive force). 
	When the Coulomb force acting on several charges, it tries to make opposite-sign charges close while like-sign charges away.
	Finally, all charges are moved to an equilibrium where all forces add up to zero, and the position of charges stays stable.
	Similarly, we can deduce that the ideal topology structure of the graph should be the same as the distribution of charges.
	
	(3) How to model attractive force and repulsive force?
	
	To address this point, we extend the concept of ``force" into relation embedding space. We define that the attractive force is a kind of calculation, which can increase the similarity between relation embeddings. On the contrary, the aim of repulsive force is to reduce the similarity between relation embeddings.
	To this end, we employ GCN to obtain information propagation between relation embeddings to play as attractive force. Besides, we use the similarity between conflicted relation embeddings as a penalty term for the objective loss function to play as repulsive force.

	\section{Method}
	
	% 这里进行task的overview
	In this section, we present our force-directed graph based distant supervised relation extraction model --- FDG-RE. We first give the task definition. Then, we provide detail formalization of the force-directed graph, with special emphasize on modeling \textit{attractive force} and \textit{repulsive force}. Finally, we introduce the implementation of relation extraction. 
	
	\subsection{Task Definition}
	
	We define relation classes as $\mathcal{R} = \{r_1, r_2, ..., r_k\}$, where $k$ is the number of relations. Given a bag of sentences $\mathcal{S}_b = \{s_1, s_2, ..., s_b\}$ consisting of $b$ sentences and an entity pair ($e_1, e_2$) presenting in all sentences. In distant supervised relation extraction, the purpose is to predict a set of target relations $\mathcal{\underline{R}}$ ($\mathcal{\underline{R}} \subseteq \mathcal{R}$) according to the entity pair ($e_1, e_2$) and the sentence-bag $\mathcal{S}_b$. Because the predicted relations often have inner connections, the goal of learning relation ties is to capture global correlation and mutual exclusion between relations.
	
	\subsection{Learning Relation Ties}
	
	This section illustrates how we construct the force-directed graph and model relation ties.
	
	% 一共存在几个关键的问题
	\subsubsection{Graph Construction}
	
\begin{figure}[h]
		\centering
		\includegraphics[width=0.99\columnwidth]{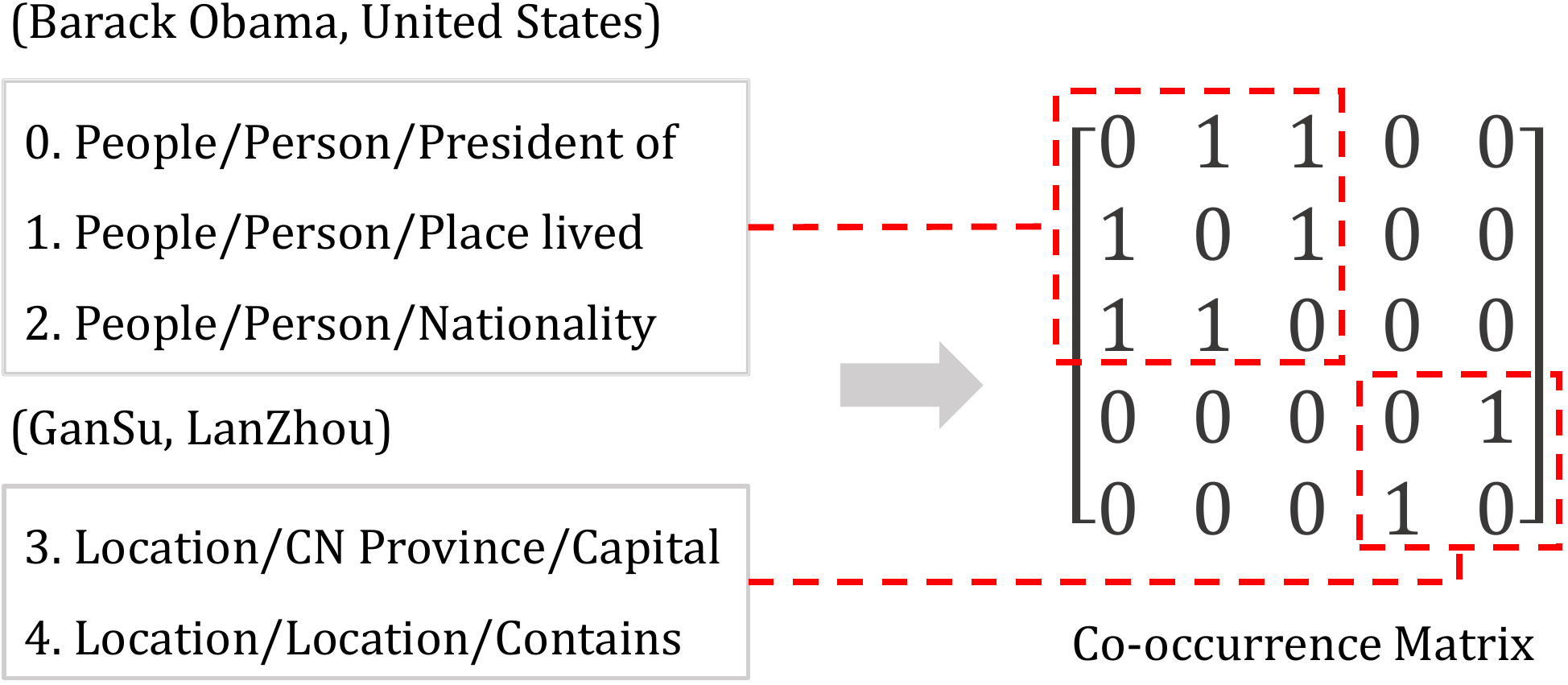} 
		\caption{An example of constructing adjacency matrix of the graph.}
		\label{fig:graph}
	\end{figure}

	In order to capture the global topology structure of relation ties, we build a graph $\mathcal{G}$ with relation embeddings as nodes $\mathcal{V} = \{\nu_1, \nu_2, ..., \nu_k\}$ and the co-occurrence between relations as edges $\mathcal{E} = \{\varepsilon_1, \varepsilon_2, ..., \varepsilon_n\}$, where $k$ and $n$ are the number of relations and edges respectively. Concretely, if two relations ($r_i, r_j$) appear in a same entity pair ($e_a, e_b$), there will be an edge $\varepsilon_h$ between the two nodes ($\nu_i, \nu_j$). Finally, as shown in Figure \ref{fig:graph}, the adjacency matrix of the graph is the symmetrical co-occurrence matrix $\mathrm{\textbf{M}}^{k\times k}$ of relations.
	
	\subsubsection{Attractive Force}
	
	In general, we conclude that the correlations between relations can have two categories: weak correlation and strong correlation. Weak correlation mainly involve the co-occurrence between relations such as \texttt{Place\_lived} and \texttt{Born\_in}. While, strong correlation means the logical entailment such as ``\texttt{President\_of} $\Rightarrow$ \texttt{Nationality}". It is worth noting that the correlation between relations is directional. For example, the probability of ``\texttt{President\_of} $\rightarrow$ \texttt{Nationality}" is 1. In contrast, the probability of ``\texttt{Nationality} $\rightarrow$ \texttt{President\_of}" is close to 0.
	To represent such asymmetric correlations, we introduce occurrence times $N_r$ of relation $r$ to the co-occurrence matrix $\mathrm{\textbf{M}}$ to get the conditional probability transition matrix $\mathrm{\textbf{P}}$:
	
	\begin{equation}
		\mathrm{\textbf{P}}_{ij} = \frac{\mathrm{\textbf{M}}_{ij}}{N_{i}},
	\end{equation}  
	
	where $\mathrm{\textbf{P}}_{ij}$ denotes the probability of ``$r_i \rightarrow r_j$".	
	As mentioned above, for an ordinary people, the probability of ``\texttt{Nationality} $\rightarrow$ \texttt{President\_of}" is close to 0. To make our model more generalizable, we filter these ``noisy transition" with a threshold $\theta$, that is, if $\mathrm{\textbf{P}}_{ij} < \theta$, $\hat{\mathrm{\textbf{P}}}_{ij} = 0$.

	% 这里还是要好好的重新说一下
	Then, we employ a $l$-layer GCN to obtain information propagation between relation embeddings:
	
	\begin{equation}
		\mathrm{\textbf{H}}^{l} = f(\hat{\mathrm{\textbf{P}}}\mathrm{\textbf{H}}^{l-1} \mathrm{\textbf{W}}^{l-1}),
	\end{equation} 
	
	where $\mathrm{\textbf{H}}^l \in \mathbb{R}^{k \times d}$ is the relation representations of the $l$-th layer and $d$ is the dimension of relation embeddings. $\hat{\mathrm{\textbf{P}}}$ denotes the filtered probability transition matrix. $\mathrm{\textbf{W}}^{l-1} \in \mathbb{R}^{d \times d}$ is the weight matrix to learn. $f(\cdot)$ is a non-linear function. 
	Intuitively, the more information is exchanged between relation embeddings, the closer their spatial locations will be.
	
	\subsubsection{Repulsive Force}
	
	To obtain global mutual exclusion information between relations, we transform co-occurrence matrix $\mathrm{\textbf{M}}$ into  mutual exclusion matrix $\mathrm{\textbf{U}}$:
	
	\begin{equation}
	\mathrm{\textbf{U}}_{ij} = \begin{cases}
	0, \qquad if \quad \mathrm{\textbf{M}}_{ij} = 1
	\\
	1, \qquad if \quad \mathrm{\textbf{M}}_{ij} = 0
	\end{cases}.
	\end{equation} 
	
	Then, we define the similarity $\xi_{ij}$ between relation embeddings $\bm{h}_i$ and $\bm{h_j}$ with a simple dot product:
	
	\begin{equation}
		\xi_{ij} = \bm{h}_i^T\bm{h_j}.
	\end{equation} 

	Note that we assume a relation is ``conflicted" with itself, that is, if $i = j, \mathrm{\textbf{U}}_{ij} = 1$. This assumption can be regarded as a $L_2$ normalization to make relation embeddings more stable.
	Thus, the global mutual exclusion between relations is defined as:
	
	\begin{equation}
		\omega = sum(\mathrm{\textbf{H}} \mathrm{\textbf{H}}^T * \mathrm{\textbf{U}}),
	\end{equation}
	
	where $*$ denotes element-wise multiply operation. Because $\omega$ is the sum of all pairwise mutual exclusion, its value is too large. We further scale it by:
	
	\begin{equation}
		\Omega = \frac{\omega}{k\times k \times d} .
	\end{equation}
	
	Finally, we leverage $\Omega$ as the penalty term for the objective loss function to act as repulsive forces.
	
	\subsection{Relation Extraction}
	
	In FDG-RE, the position embeddings proposed by \citet{zeng2014relation} are adopted to specify the target entity pair ($e_1, e_2$) and make model pay more attention to the words close to target entities. The final representation of a word $\bm{v_i}$ is the concatenation of word embedding $\bm{w}_i$ and two position embeddings $\bm{p}_i^1, \bm{p}_i^2$: 
	
	\begin{equation}
		\bm{v}_i = [\bm{w}_i;\bm{p}_i^1;\bm{p}_i^2].
	\end{equation}
	
	We employ PCNN~\cite{zeng2015distant} to learn sentence-level features, which mainly consists of two parts: a traditional convolutional neural network (CNN) and piece-wise max-pooling. Suppose $\mathrm{\textbf{C}}_i$ is one of the feature maps learned by CNN, PCNN divides every feature map $\mathrm{\textbf{C}}_i$ into three parts \{ $\mathrm{\textbf{C}}_{i1}, \mathrm{\textbf{C}}_{i2}, \mathrm{\textbf{C}}_{i3}$\} by the position of two target entities ($e_1, e_2$). Then, the max-pooling operation is performed on the three parts separately. The final sentence representation $\bm{s}_i$ is the concatenation of all vectors:
	
	\begin{equation}
		\bm{s}_i = [\mathrm{\textbf{C}}_{i1};\mathrm{\textbf{C}}_{i2};\mathrm{\textbf{C}}_{i3}],
	\end{equation}
	
	We employ sentence-level selective attention to combine embedded sentences into one bag representation $\bm{b}_i$, aiming to aggregate information across sentences:
	
	\begin{equation}
		\bm{b}_i = \sum_{1}^{b}  \alpha_j \bm{s}_j,
	\end{equation}
	
	where $\alpha_j$ is calculated by:
	
	\begin{equation}
		\alpha_j = \frac{exp(\bm{s}_j^T \bm{h}_r)}{\sum_{1}^{b} exp(\bm{s}_j^T \bm{h}_r)}.
	\end{equation}
	
	$\alpha_j$ is a coupling coefficient which scores how well the input sentence $s_j$ and the target relation $r$ matches. 
	The output of the neural network is:
	
	\begin{equation}
		\mathrm{\textbf{O}} = \mathrm{\textbf{B}}^T \mathrm{\textbf{H}}
	\end{equation}
	
	where $\mathrm{\textbf{B}}$ denotes the sentence-bag representation matrix, $\mathrm{\textbf{H}}$ is the relation representations learned by GCN, the bias term in this equation is omitted for convenient description. In fact, $\mathrm{\textbf{H}}$ is an inner-dependent classification network which has learned the correlations between relations. We employ softmax to get the final prediction probability:
	
	\begin{equation}
		pred(r | \mathrm{\textbf{B}}_i, \boldsymbol{\Theta}) = \dfrac {exp(\bm{o}_r)}{\sum_{j=1}^k exp(\bm{o}_j)}
	\end{equation}
	
	Finally, the objective function of the model is:
	
	\begin{equation}
		\mathcal{L}\left( \boldsymbol{\Theta} \right) = - \sum_{i=1}^{n} log p(r_i,|\mathrm{\textbf{B}}_i, \boldsymbol{\Theta}) + \lambda \Omega.
	\end{equation}
	
	Where $\Omega$ represents the repulsive forces between mutual exclusive relations obtained by equation (6), $\lambda$ is a harmonic factor that balances the two terms. $r_i$ is the predicted relations of sentence-bag $\mathrm{\textbf{B}}_i$. $\boldsymbol\Theta$ indicates all parameters of the model.
	
	\section{Experiments}
	
	\begin{figure*}[t]
		\centering
		\subfigure[PCNN+ATT]{
			\begin{minipage}[b]{0.66\columnwidth}
				\includegraphics[width=1\columnwidth]{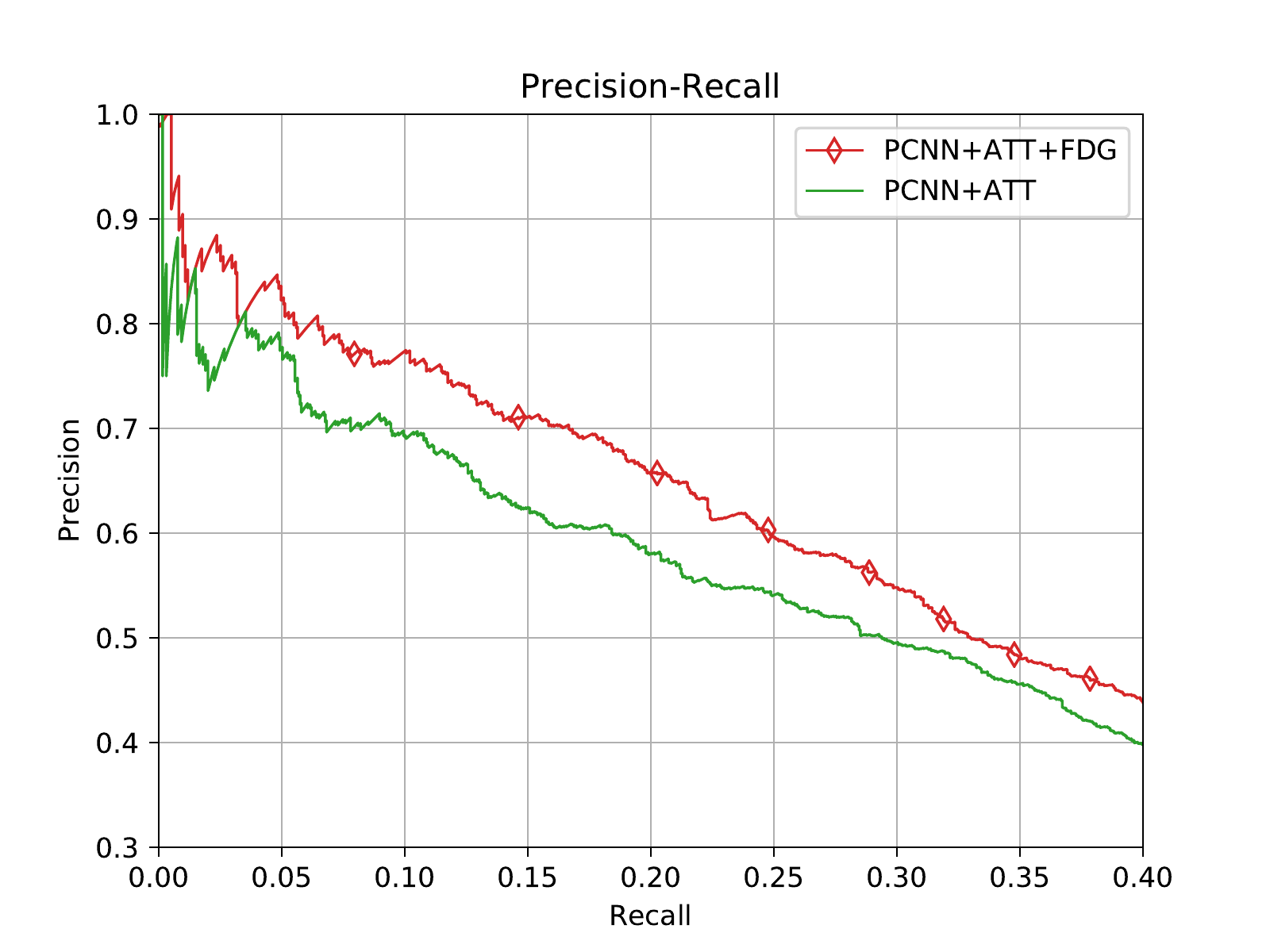}
		\end{minipage}}
		\subfigure[PCNN+AVE]{
			\begin{minipage}[b]{0.66\columnwidth}
				\includegraphics[width=1\columnwidth]{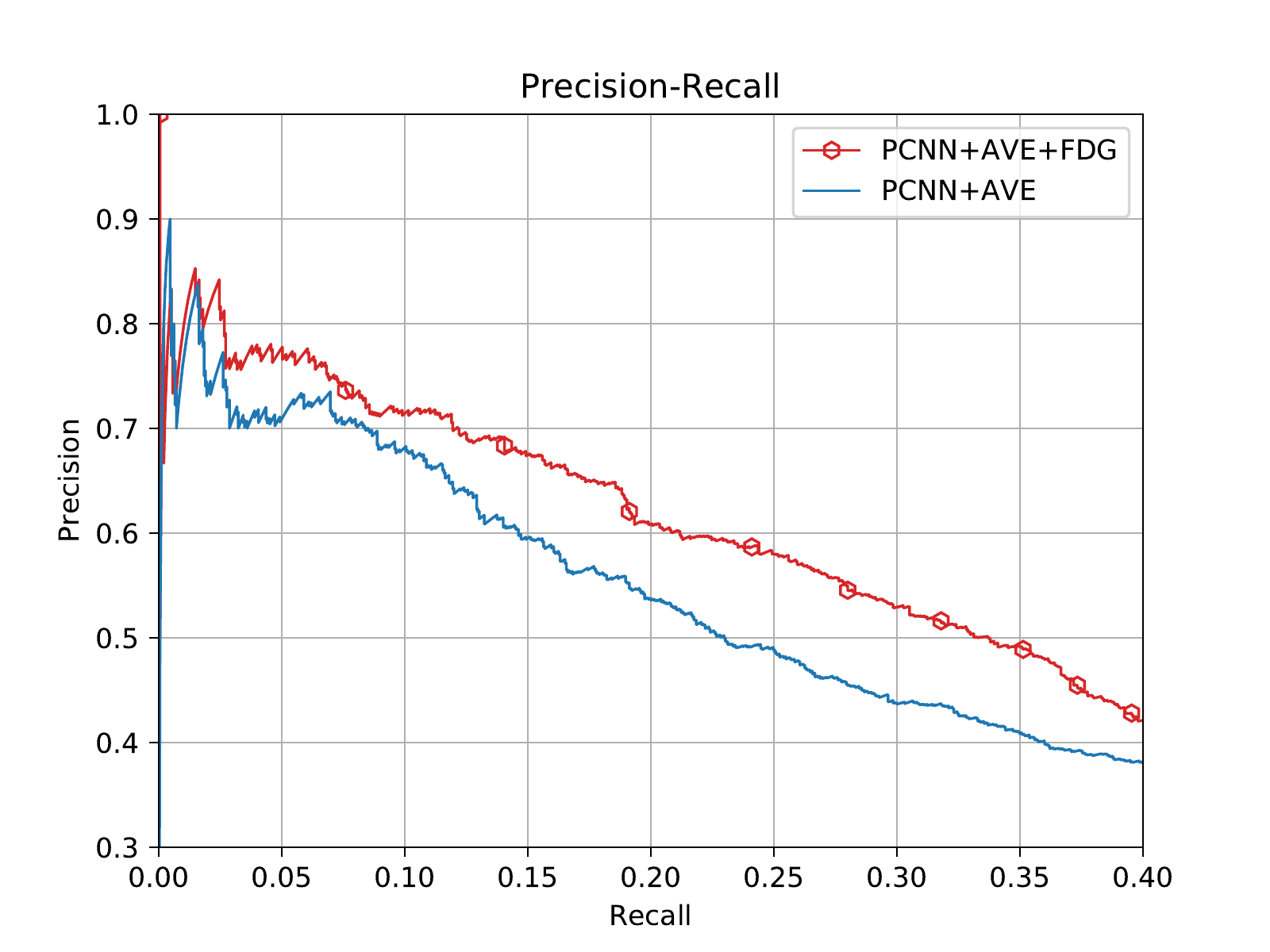}
		\end{minipage}}
		\subfigure[PCNN+ONE]{
			\begin{minipage}[b]{0.66\columnwidth}
				\includegraphics[width=1\columnwidth]{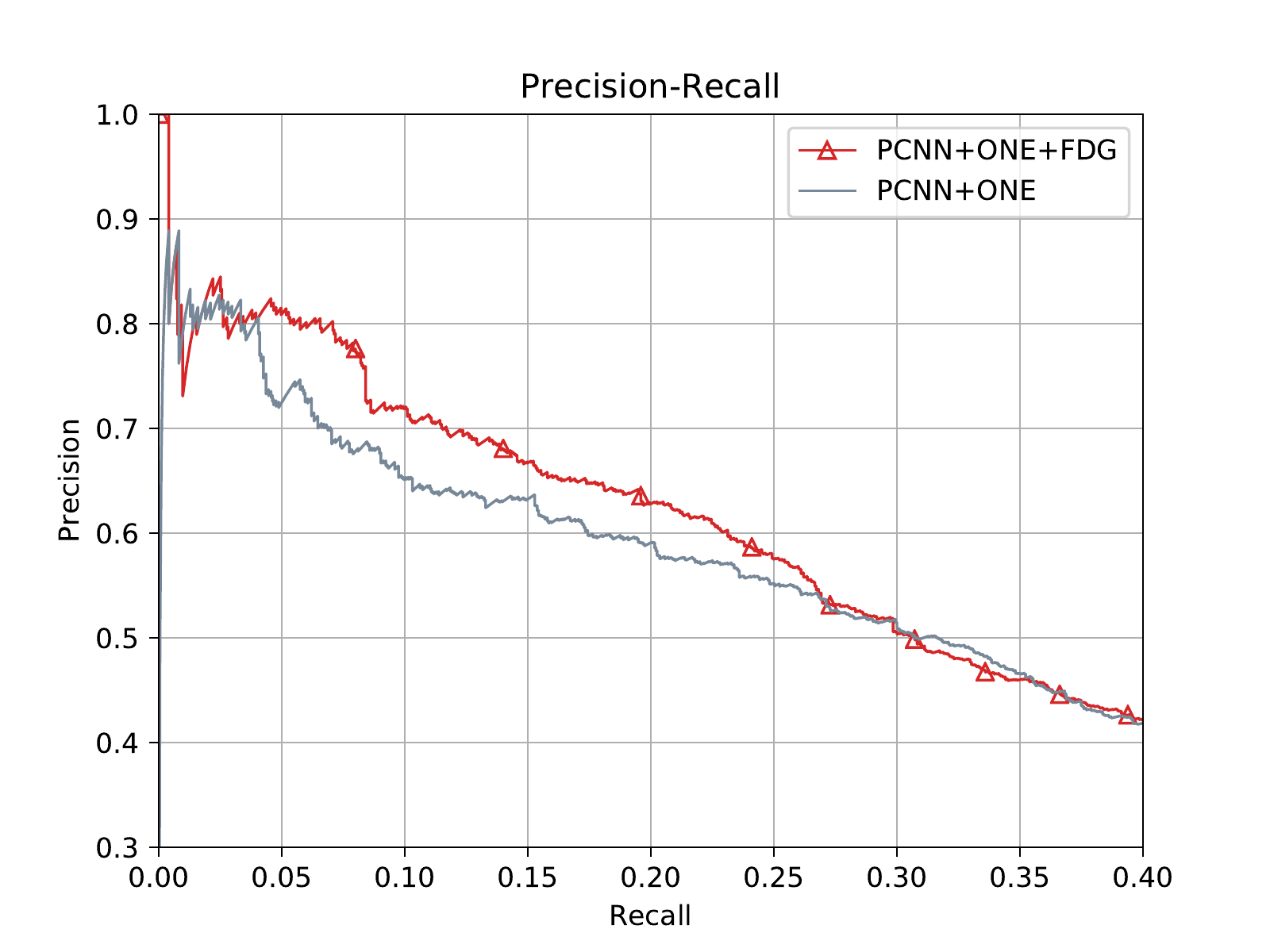}
		\end{minipage}}
		\caption{The proposed force-directed graph is applied as a module to augment three different relation extraction methods.}
		\label{fig:module}
	\end{figure*}
	
	Our experiments are designed to demonstrate four points:
	
	\begin{enumerate}
		\item The proposed force-directed graph can be used as a module to augment existing relation extraction methods and significantly improve their performance (section \ref{point1}).
		
		\item Among the similar methods of learning relation ties, our FDG-RE performs best (section \ref{point2}).
		
		\item FDG-RE outperforms the state-of-the-art distant supervised relation extraction methods (section \ref{point3}).
		
		\item FDG-RE can indeed learn the topology structure of relation ties (section \ref{point4}).
	\end{enumerate}

	In the following, we first introduce the dataset and evaluation metrics. Second, we show the experimental setup. Third, we conduct three parts of detailed comparison in response to the experimental purposes 1-3. Finally, we illustrate the visualization of relation embeddings in response to the experimental purpose 4.
	
	\subsection{Dataset and Evaluation Metrics}
	
	We evaluate our FDG-RE and all baselines on a widely used dataset NYT developed by \citet{riedel2010modeling}, which was structured by aligning relations in Freebase \cite{bollacker2008freebase} with the New York Times (NYT) corpus. In the NYT dataset, training sentences are from 2005-2006 corpus and test sentences are from 2007. Specifically, it contains 520K training sentences and 172K test sentences. There are 53 unique relations including a special relation \textsl{NA} that signifies no relation between the entity pair.
	
	Following the previous methods \cite{zeng2015distant,lin2016neural,jointly,su2018exploring,shang2020are}, we evaluate all models in held-out evaluation and present precision-recall curves (PR-Curves). The held-out evaluation is an approximate measure of the model, which uses the extracted relations to automatically compare with the fact in knowledge graph.
	
	\subsection{Setup}
	
	For all baselines, during training, we follow the settings used in their papers. We set the hyper-parameters in FDG-RE by Random Search \cite{bergstra2012random}. Table \ref{tab:set} shows the parameters used in FDG-RE.
	
 	\begin{table}[h]
	 	\centering
	 	\begin{tabular}{lr}  
	 		\toprule
	 		Setting  & Number \\
	 		\midrule
	 		Kernel size &   3\\
	 		Number of feature maps & 320  \\
	 		Word embedding dimension & 50 \\
	 		Position embedding dimension& 5\\
	 		learning rate & 0.19\\
	 		Threshold $\theta$ & 0.18\\
	 		Harmonic factor $\lambda$ & 0.25\\
	 		Number of GCN layers $l$ & 2\\
	 		\bottomrule
	 	\end{tabular}
	 	\caption{Parameters Setting}
	 	\label{tab:set}
	 \end{table}
	
	\subsection{Act as a Module}
	\label{point1}
	
	In this section, we conduct experiments to demonstrate that the proposed force-directed graph can be applied as a module to augment existing relation extraction methods and significantly improve their performance.
	
	\subsubsection{Baselines}
	
	We select three conventional relation extraction methods as baselines. During extraction, they all predict relations independently and ignore the relation ties.
	
	\begin{itemize}
		\item \textbf{PCNN+ATT}: \citet{lin2016neural} propose to use sentence-level attention mechanism to obtain sentence-bag representations.
		
		\item \textbf{PCNN+AVE}: We try to obtain the bag-level representations via the average of all the sentence representations.
		
		\item \textbf{PCNN+ONE}: \citet{zeng2015distant} propose to use the feature of the most correct sentence to represent the sentence-bag.
		
	\end{itemize}

	\subsubsection{Results}

	The results are shown in Figure \ref{fig:module}, \textbf{+FDG} means applying the force-directed graph to the corresponding model. It can be observed that the proposed module can significantly improve the performance of three baselines. This proves that: (1) Considering relation ties in distant supervised relation extraction can indeed reduce the potential searching space and improve the prediction performance. (2) The proposed force-directed graph is flexible and adaptable.
	
	\subsection{Compare with Similar Methods}
	\label{point2}
	
	In this part, we compare FDG-RE with similar methods which focus on learning relation ties to show that our model performs best.
	
	\subsubsection{Baselines}
	
	We use the following four models as baselines:
	
	\begin{itemize}
		\item \textbf{MIMLCNN}: \citet{jiang2016relation} obtain relation dependencies by designing multi-label loss function in the neural network classifier.
		
		\item \textbf{Rank+ExATT}: \citet{jointly} adopt pairwise learning to rank framework to capture the co-occurrence dependency between relations.
		
		\item \textbf{Memory}: \citet{feng2017effective} use memory network to capture relation dependencies.
		
		\item \textbf{PartialMax+IQ+ATT}: \citet{su2018exploring} utilize the Encoder-Decoder framework to capture relation dependencies and predict relations with a RNN decoder.
	\end{itemize}
	
	 We implemented MIMLCNN and PartialMax+IQ+ATT. For Rank+ExATT\footnote{https://github.com/oceanypt/DR\_RE} and Memory\footnote{https://github.com/liuyongjie985/Effective\_Deep\_Memory\_Net- works\_for\_Distant\_Supervised\_Relation\_Extraction}, we use the codes provided by authors.
	
	\subsubsection{Results}
	
	\begin{figure}[t]
		\centering
		\includegraphics[width=0.9\columnwidth]{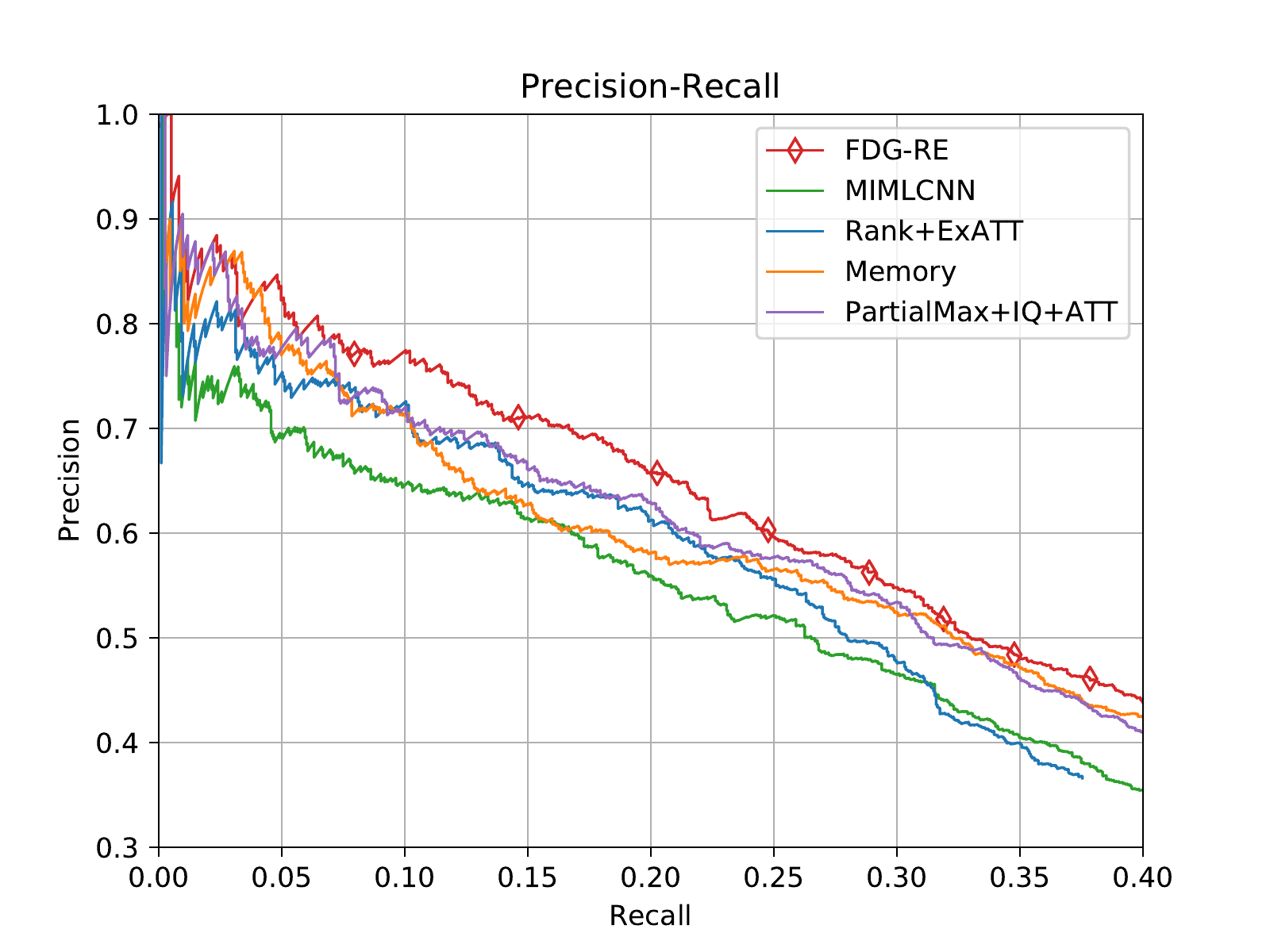} 
		\caption{Comparison with the similar methods.}
		\label{fig:relationties}
	\end{figure}

	Figure \ref{fig:relationties} shows the resulting PR-Curves in the most concerned area. It can be observed that:
	% 说显式方法和隐式方法
	
	(1) Comparing the explicit methods (FDG-RE, PartialMax+IQ+ATT) with the implicit methods (MIMLCNN, Rank+ExATT, Memory), we can conclude that the explicit methods perform better than implicit methods. Because RNN can well describe the linear dependencies between relations, and GCN is good at learning the regional dependencies. In other words, using RNN or GCN means the prior knowledge of the topology structure is added at the beginning of the training process.   
	% 说我们的方法和Encoder-Decoder之间的对比
	
	(2) Among the two explicit methods, FDG-RE can consistently and significantly outperform PartialMax+IQ+ATT in the entire range of recall. This proves that considering global connections is better than focusing on local dependencies. Concretely, PartialMax+IQ+ATT tries to use a linear Encoder-Decoder framework to learn relation ties. However, the pre-defined order of relations has an important influence on the final prediction. For example, if decoder has predicted relation \texttt{Nationality}, it will not predict \texttt{President\_of} at later steps. As a result, all the downstream relations of \texttt{President\_of} will be unreachable.
	
	Overall, the experimental results demonstrate that the motivation of our work that using force-directed graph to model global topology structure of relation ties is effective.
	
	\subsection{Compare with SOTA Methods}
	\label{point3}
	
	We further compare FDG-RE with the latest distant supervised relation extraction methods to illustrate that our model achieves the state-of-the-art performance.
	
	\subsubsection{Baselines}
	
	Our model does not use any external information (e.g., entity type, entity description and so on). Therefore, we select three latest methods that do not use external information as baselines:
	
	\begin{itemize}
		\item \textbf{PCNN+C2SA}: \citet{yuan2019cross} use cross-relation cross-bag selective attention to deal with the noisy labeling problem.
		
		\item \textbf{PCNN+ATT\_RA+BAG\_ATT}: \citet{ye-ling-2019-distant} propose intra-bag and inter-bag attention to alleviate the influence of noisy sentences.
		
		\item \textbf{DCRE}: \cite{shang2020are} try to convert noisy sentences into useful training instances by unsupervised deep clustering.
	\end{itemize}
	
	 We implemented DCRE, for  PCNN+C2SA\footnote{https://github.com/yuanyu255/PCNN\_C2SA} and PCNN+ATT\_RA+BAG\_ATT\footnote{https://github.com/ZhixiuYe/Intra-Bag-and-Inter-Bag-Attentions}, we use the codes provided by authors. Their original paper use the NYT set which contains 570k training sentences, whose training set contains lots of test fact. Following the mainstream of distant supervised relation extraction, in our experiments, they are evaluated by the filtered NYT\footnote{https://github.com/thunlp/NRE} set which has 520k training sentences.
	
	\subsubsection{Results}
\begin{figure}[t]
		\centering
		\includegraphics[width=0.9\columnwidth]{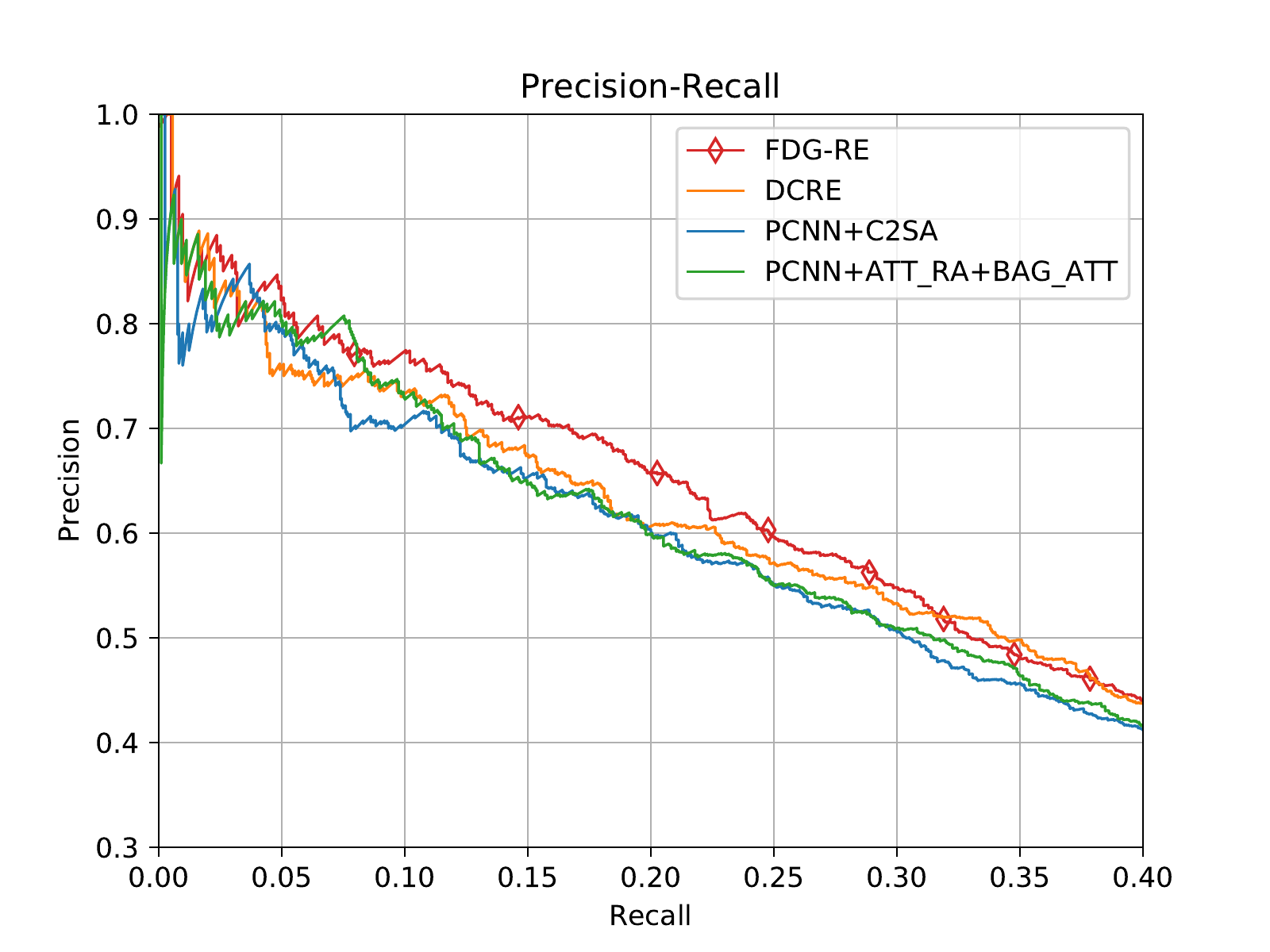} 
		\caption{Comparison with the state-of-the-art methods.}
		\label{fig:soat}
	\end{figure}

	As shown in Figure \ref{fig:soat}, there is an obvious margin between FDG-RE and the three baselines. We believe that this observation is mainly due to (1) The three baselines all predict relations independently and ignore the relation ties. On the contrary, FDG-RE considers global correlation and mutual exclusion between relations. (2) The objective function of FDG-RE has loss term $\sum_{i=1}^{n} logp(r_i,|\mathrm{\textbf{B}}_i, \boldsymbol{\Theta})$ and penalty term $\lambda \Omega$. It can not only penalize the false classifications, but also enhance the generalization ability of the model. 
	
	\subsection{Topology Structure of Relation Ties}
	\label{point4}
	
\begin{figure}[t]
		\centering
		\subfigure[FDG-RE]{
			\begin{minipage}[b]{0.99\columnwidth}
				\includegraphics[width=1\columnwidth]{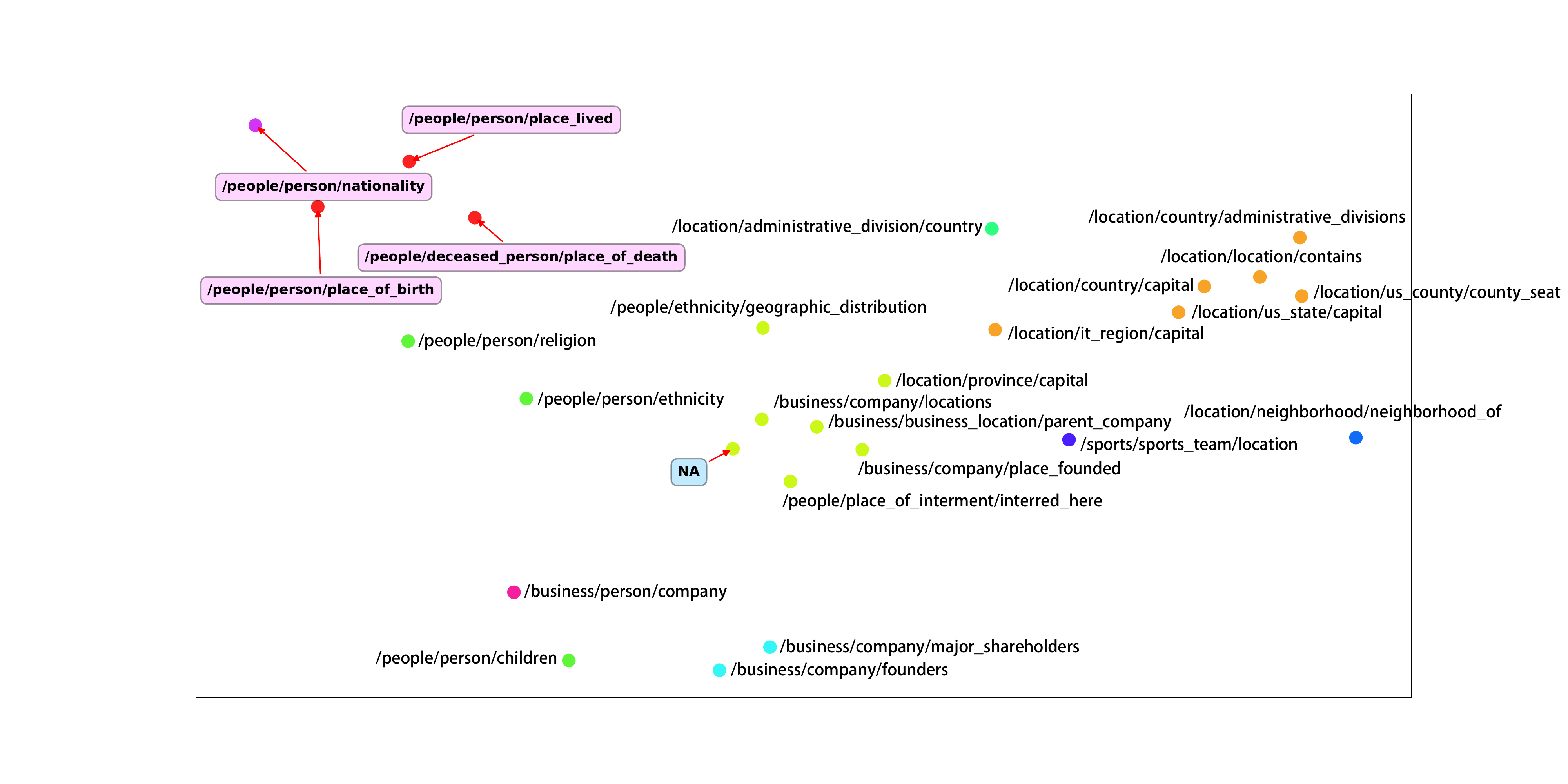}
		\end{minipage}}
		\subfigure[PCNN+ATT]{
			\begin{minipage}[b]{0.99\columnwidth}
				\includegraphics[width=1\columnwidth]{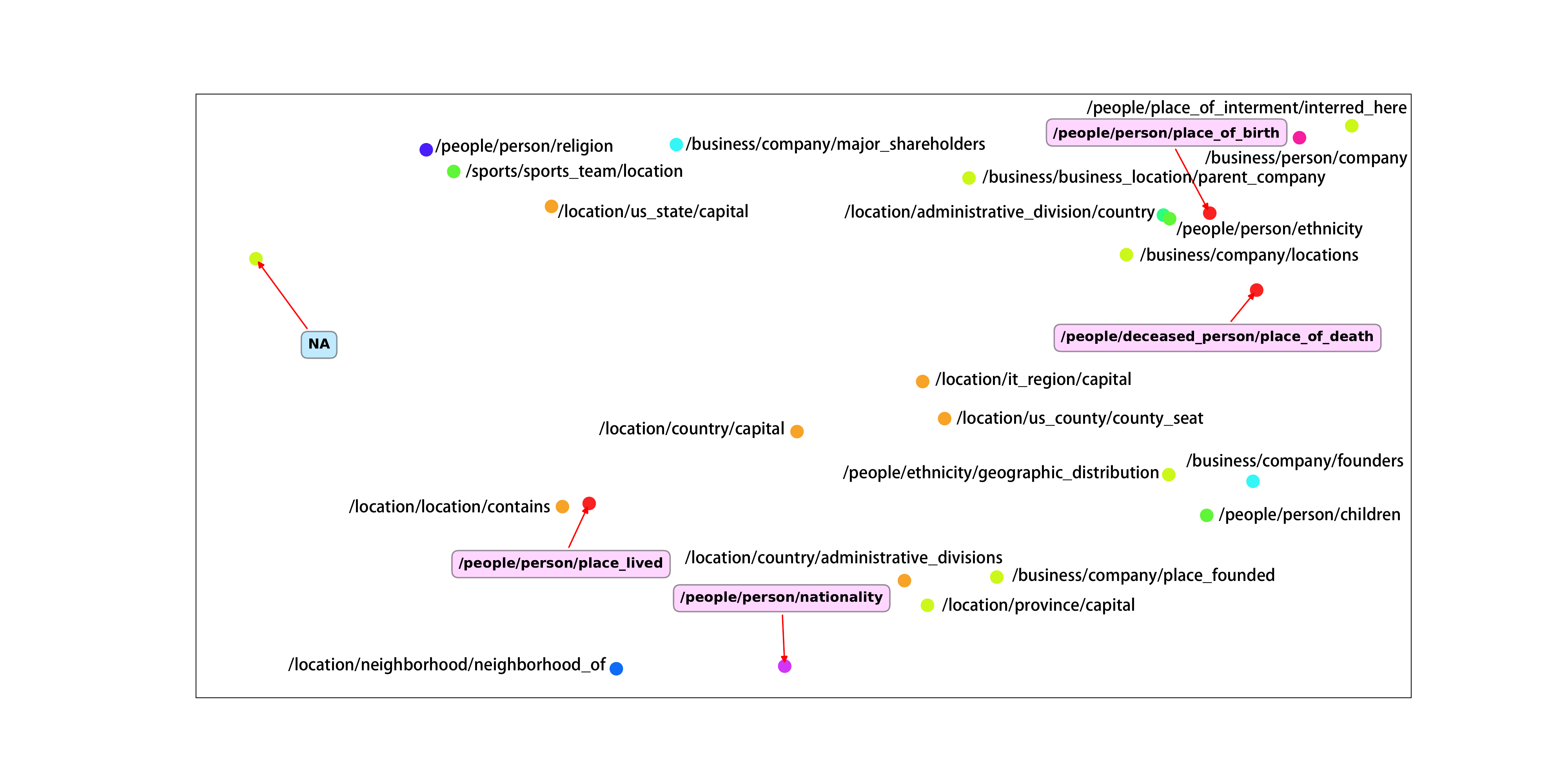}
		\end{minipage}}
		\caption{The visualization of relation embeddings learned by FDG-RE and PCNN+ATT.}
		\label{fig:vialization}
	\end{figure}

	In order to prove that our proposed method can indeed obtain the global topology structure of relation ties, we visualize the relation embeddings learned by FDG-RE with Isomap \cite{tenenbaum2000global}, which is a nonlinear dimensionality reduction algorithm. The relation embeddings learned by PCNN+ATT are also visualized as a comparison. The results are shown in Figure \ref{fig:vialization}. During visualization, we omit the long-tail relations to highlight key information.
	It can be observed that: 
	
	(1) The topology structure of relation ties learned by FDG-RE is inter-compact and intro-loose, which shows the characteristics of clustering. 
	For example, the position of four connected relations \texttt{Nationality}, \texttt{Place\_lived}, \texttt{Place\_of\_birth} and \texttt{Place\_of\_death} are close to each other. While, they are far away from the relations whose root node is \texttt{Location} or \texttt{Business}. This is consistent with our motivation. 
	In contrast, the relation embeddings learned by PCNN+ATT are almost randomly distributed.  
	
	(2) The relation \textsl{NA} is in the center of Figure \ref{fig:vialization} (a). It means ``no relations " and is conflicted with all the other relations. Because the effect of repulsive forces, it ``pushes" other relations away. However, PCNN cannot obtain such features. 
	
	(3) To a certain extent, the relation representations learned by our force-directed graph maintain some ``semantic" correlations. For example, the relations in the same branch \texttt{Location} are close to each other. Therefore, the relation embeddings have the generalization abilities when perform as a relation classifier. This proves the experimental conclusion of section \ref{point1} from the side.

	\section{Related Work}
	
	An entity pair may have multiple relations in knowledge graph. Therefore, previous studies formalize distant supervised relation extraction as a multi-instance multi-label prediction task \cite{riedel2010modeling,Hoffmann2011Knowledge}.
	Afterwards, there are many attempts focusing on exploring the correlation and mutual exclusion between relations to reduce the potential searching space.  Existing approaches can be broadly divided into two categories: 
	
	The first category explicitly represents relation ties by the model architecture. For example,
	\citet{han2016global} try to utilize Markov Logic Network to represent the transition probability between relations.
	\citet{su2018exploring} employ Encoder-Decoder framework to capture relation connections.
	While benefiting from the model architecture, these methods are limited by the learning ability of the model.  
	Specifically, Markov Logical Network can only consider the neighboring nodes (Markov property), and Encoder-Decoder framework learns relation dependencies in a linear manner.
	As a consequence, they can only obtain a range of relation ties, rather than global connections between relations.
	
	The second category implicitly captures relation ties by soft constraints. For example,
	\citet{jiang2016relation} handle relation connections by using a shared entity-pair-level representation and designing multi-label classification loss function. 
	\citet{jointly} adopt pairwise learning to rank framework to capture co-occurrence dependencies among relations. 
	\citet{feng2017effective} propose a two-layer memory network and employ the attention mechanism to learn relation dependencies.
	However, these methods greedily capture relation ties according to the current sentence-bag. Although the training process will traverse all sentence-bags, focusing on local features can not yield a precise global topology structure of relation ties.
	
	Different from the aforementioned two kinds of methods, the model proposed in this paper explicitly learns relation correlations by using GCN to obtain information propagation based on a directed graph, and implicitly captures mutual exclusion by adding penalty term into the objective loss function. 
	Experimental results demonstrate that our proposed force-directed graph can indeed capture the global topology structure of relation ties, and can be used as a module to augment existing relation extraction methods.
	
	\section{Conclusion}
	In this paper, we study learning relation ties in distant supervised relation extraction and propose a novel force-directed graph based relation extraction model, named FDG-RE. 
	As compared with previous methods, our FDG-RE introduces the concept of attractive force and repulsive force into relation embedding space and can indeed capture the global topology structure of relation ties.
	We conduct various experiments on a widely used benchmark dataset and the evaluation results show that our model outperforms state-of-the-art baselines. 
	Besides, the proposed force-directed graph is flexible and adaptable, it can be used as a module to augment other relation extraction methods.
	
	\bibliographystyle{named}
	\bibliography{ijcai20}

\begin{thebibliography}{}

\bibitem[\protect\citeauthoryear{Bergstra and
  Bengio}{2012}]{bergstra2012random}
James Bergstra and Yoshua Bengio.
\newblock Random search for hyper-parameter optimization.
\newblock {\em Journal of Machine Learning Research}, 13(Feb):281--305, 2012.

\bibitem[\protect\citeauthoryear{Bollacker \bgroup \em et al.\egroup
  }{2008}]{bollacker2008freebase}
Kurt Bollacker, Colin Evans, Praveen Paritosh, Tim Sturge, and Jamie Taylor.
\newblock Freebase: a collaboratively created graph database for structuring
  human knowledge.
\newblock In {\em Proceedings of the 2008 ACM SIGMOD international conference
  on Management of data}, pages 1247--1250. AcM, 2008.

\bibitem[\protect\citeauthoryear{Feng \bgroup \em et al.\egroup
  }{2017}]{feng2017effective}
Xiaocheng Feng, Jiang Guo, Bing Qin, Ting Liu, and Yongjie Liu.
\newblock Effective deep memory networks for distant supervised relation
  extraction.
\newblock In {\em IJCAI}, pages 4002--4008, 2017.

\bibitem[\protect\citeauthoryear{Halliday \bgroup \em et al.\egroup
  }{2013}]{halliday2013fundamentals}
David Halliday, Robert Resnick, and Jearl Walker.
\newblock {\em Fundamentals of physics}.
\newblock John Wiley \& Sons, 2013.

\bibitem[\protect\citeauthoryear{Han and Sun}{2016}]{han2016global}
Xianpei Han and Le~Sun.
\newblock Global distant supervision for relation extraction.
\newblock In {\em Thirtieth AAAI Conference on Artificial Intelligence}, 2016.

\bibitem[\protect\citeauthoryear{Hoffmann \bgroup \em et al.\egroup
  }{2011}]{Hoffmann2011Knowledge}
Raphael Hoffmann, Congle Zhang, Xiao Ling, Luke~S. Zettlemoyer, and Daniel~S.
  Weld.
\newblock Knowledge-based weak supervision for information extraction of
  overlapping relations.
\newblock In {\em Meeting of the Association for Computational Linguistics:
  Human Language Technologies}, 2011.

\bibitem[\protect\citeauthoryear{Jiang \bgroup \em et al.\egroup
  }{2016}]{jiang2016relation}
Xiaotian Jiang, Quan Wang, Peng Li, and Bin Wang.
\newblock Relation extraction with multi-instance multi-label convolutional
  neural networks.
\newblock In {\em Proceedings of COLING 2016, the 26th International Conference
  on Computational Linguistics: Technical Papers}, pages 1471--1480, 2016.

\bibitem[\protect\citeauthoryear{Kipf and Welling}{2017}]{kipf2017semi}
Thomas~N. Kipf and Max Welling.
\newblock Semi-supervised classification with graph convolutional networks.
\newblock In {\em International Conference on Learning Representations (ICLR)},
  2017.

\bibitem[\protect\citeauthoryear{Lin \bgroup \em et al.\egroup
  }{2016}]{lin2016neural}
Yankai Lin, Shiqi Shen, Zhiyuan Liu, Huanbo Luan, and Maosong Sun.
\newblock Neural relation extraction with selective attention over instances.
\newblock In {\em Proceedings of the 54th Annual Meeting of the Association for
  Computational Linguistics (Volume 1: Long Papers)}, volume~1, pages
  2124--2133, 2016.

\bibitem[\protect\citeauthoryear{Mintz \bgroup \em et al.\egroup
  }{2009}]{mintz2009distant}
Mike Mintz, Steven Bills, Rion Snow, and Dan Jurafsky.
\newblock Distant supervision for relation extraction without labeled data.
\newblock In {\em Proceedings of the Joint Conference of the 47th Annual
  Meeting of the ACL and the 4th International Joint Conference on Natural
  Language Processing of the AFNLP: Volume 2-Volume 2}, pages 1003--1011.
  Association for Computational Linguistics, 2009.

\bibitem[\protect\citeauthoryear{Riedel \bgroup \em et al.\egroup
  }{2010}]{riedel2010modeling}
Sebastian Riedel, Limin Yao, and Andrew McCallum.
\newblock Modeling relations and their mentions without labeled text.
\newblock In {\em Joint European Conference on Machine Learning and Knowledge
  Discovery in Databases}, pages 148--163. Springer, 2010.

\bibitem[\protect\citeauthoryear{Shang \bgroup \em et al.\egroup
  }{2020}]{shang2020are}
Yu~Ming Shang, Heyan Huang, Xianling Mao, Xin Sun, and Wei Wei.
\newblock Are noisy sentences useless for distant supervised relation
  extraction.
\newblock In {\em Proceedings of AAAI}, 2020.

\bibitem[\protect\citeauthoryear{Su \bgroup \em et al.\egroup
  }{2018}]{su2018exploring}
Sen Su, Ningning Jia, Xiang Cheng, Shuguang Zhu, and Ruiping Li.
\newblock Exploring encoder-decoder model for distant supervised relation
  extraction.
\newblock pages 4389--4395, 2018.

\bibitem[\protect\citeauthoryear{Tenenbaum \bgroup \em et al.\egroup
  }{2000}]{tenenbaum2000global}
Joshua~B Tenenbaum, Vin De~Silva, and John~C Langford.
\newblock A global geometric framework for nonlinear dimensionality reduction.
\newblock {\em science}, 290(5500):2319--2323, 2000.

\bibitem[\protect\citeauthoryear{Ye and Ling}{2019}]{ye-ling-2019-distant}
Zhi-Xiu Ye and Zhen-Hua Ling.
\newblock Distant supervision relation extraction with intra-bag and inter-bag
  attentions.
\newblock In {\em Proceedings of the 2019 Conference of the North {A}merican
  Chapter of the Association for Computational Linguistics: Human Language
  Technologies, Volume 1 (Long and Short Papers)}, pages 2810--2819,
  Minneapolis, Minnesota, 2019. Association for Computational Linguistics.

\bibitem[\protect\citeauthoryear{Ye \bgroup \em et al.\egroup }{2017}]{jointly}
Hai Ye, Wenhan Chao, Zhunchen Luo, and Zhoujun Li.
\newblock Jointly extracting relations with class ties via effective deep
  ranking.
\newblock In {\em Proceedings of the 55th Annual Meeting of the Association for
  Computational Linguistics (Volume 1: Long Papers)}. Association for
  Computational Linguistics, 2017.

\bibitem[\protect\citeauthoryear{Yuan \bgroup \em et al.\egroup
  }{2019}]{yuan2019cross}
Yujin Yuan, Liyuan Liu, Siliang Tang, Zhongfei Zhang, Yueting Zhuang, Shiliang
  Pu, Fei Wu, and Xiang Ren.
\newblock Cross-relation cross-bag attention for distantly-supervised relation
  extraction.
\newblock In {\em Proceedings of the AAAI Conference on Artificial
  Intelligence}, volume~33, pages 419--426, 2019.

\bibitem[\protect\citeauthoryear{Zeng \bgroup \em et al.\egroup
  }{2014}]{zeng2014relation}
Daojian Zeng, Kang Liu, Siwei Lai, Guangyou Zhou, and Jun Zhao.
\newblock Relation classification via convolutional deep neural network.
\newblock In {\em Proceedings of COLING 2014, the 25th International Conference
  on Computational Linguistics: Technical Papers}, pages 2335--2344, 2014.

\bibitem[\protect\citeauthoryear{Zeng \bgroup \em et al.\egroup
  }{2015}]{zeng2015distant}
Daojian Zeng, Kang Liu, Yubo Chen, and Jun Zhao.
\newblock Distant supervision for relation extraction via piecewise
  convolutional neural networks.
\newblock In {\em Proceedings of the 2015 Conference on Empirical Methods in
  Natural Language Processing}, pages 1753--1762, 2015.

\end{thebibliography}

\end{document}